\documentclass[11pt,a4paper]{llncs}
\usepackage{amsmath}
\usepackage{amssymb}
\usepackage{graphicx}
\usepackage{marvosym}
\usepackage{url}
\usepackage{fancyhdr}

\usepackage{geometry}
\newcommand{\keywords}[1]{\par\addvspace\baselineskip
\noindent\keywordname\enspace\ignorespaces#1}

\fancypagestyle{firstpage}{\fancyhf{}
}

\usepackage{makecell}
\usepackage{array}

\begin{document}


\title{\LARGE{MeVe: A Modular System for Memory Verification and Effective Context Control in Language Models}}

\author{\large{Andreas Ottem}}
\institute{\large{Independent Researcher \\ Alta, Norway}}

\maketitle

\thispagestyle{firstpage}

\begin{abstract}
Retrieval-Augmented Generation (RAG) systems typically face constraints because of their inherent mechanism: a simple top-k semantic search \cite{5}. The approach often leads to the incorporation of irrelevant or redundant information in the context, degrading performance and efficiency \cite{1}\cite{2}. This paper presents MeVe, a novel modular architecture intended for Memory Verification and smart context composition.

MeVe rethinks the RAG paradigm by proposing a five-phase modular design that distinctly breaks down the retrieval and context composition process into distinct, auditable, and independently tunable phases: initial retrieval, relevance verification, fallback retrieval, context prioritization, and token budgeting. This architecture enables fine-grained control of what knowledge is made available to an LLM, enabling task-dependent filtering and adaptation. We release a reference implementation of MeVe as a proof of concept and evaluate its performance on knowledge-heavy QA tasks over a subset of English Wikipedia \cite{35}. Our results demonstrate that by actively verifying information before composition, MeVe significantly improves context efficiency, achieving a 57\% reduction on the Wikipedia dataset and a 75\% reduction on the more complex HotpotQA dataset compared to standard RAG implementations \cite{37}. This work provides a framework for more scalable and reliable LLM applications. By refining and distilling contextual information, MeVe offers a path toward better grounding and more accurate factual support \cite{4}.

\end{abstract}

\section{Introduction}
The impressive abilities of Large Language Models (LLMs) are fundamentally reliant on the ability to reason in a particular context \cite{6} \cite{7}. As models continue to utilize larger context windows and retrieve information from external memory, the key challenge shifts from simple information retrieval to smart contextual management \cite{8}. It includes the dynamic fetching of contextually relevant knowledge, rigorous verification of its relevance, and efficient composition with the objective of improving LLM performance \cite{6} \cite{10}.

Retrieval-Augmented Generation (RAG) has also become a strong baseline for augmenting LLMs with out-of-model knowledge \cite{5}.

Traditional RAG systems commonly operate by performing a direct, often "top-k" semantic search against a pre-indexed knowledge base, retrieving the most similar documents and directly appending them to the LLMs input prompt. While effective in providing external knowledge, this straightforward mechanism can lead to issues such as including irrelevant, redundant or even contradictory information \cite{10} \cite{5} \cite{3}. Yet, typical implementations tightly intertwine retrieval and injection into a single monolithic step, usually by picking a fixed number of embedding-nearest neighbors and simply adding them to the prompt \cite{3} \cite{12}. This common practice makes verification implicit, heavily reduces modularity, and tends to have a difficult time gracefully scaling to dynamic memory expansion or tight token budget requirements \cite{14}. The direct injection of possibly irrelevant or redundant information can lead to "context pollution" \cite{2} \cite{10}, degrading the quality and effectiveness of LLM responses and increasing the risk of factual error or "hallucinations” \cite{1} \cite{10} \cite{15}. This challenge is compounded by the fact that retrieval quality is highly dependent on the preciseness and structure of the knowledge corpus itself \cite{3}.

As a solution to such fundamental shortcomings, we present MeVe, a new method that fundamentally redesigns retrieval, verification, prioritization, and budgeting as separate, configurable, and auditable phases \cite{16} \cite{17}. MeVe's key contribution is its modular decomposition of the RAG pipeline, transforming it from a singular operation into a structured, multi-stage process with explicit control over content quality and composition. MeVe does not aim to substitute RAG but instead redefines traditional RAG as one possible configuration within a more sophisticated, memory-aware architecture. The modular design founded upon the principles of explicit verification of information and strict retrieval mechanisms aim to alleviate typical failure modes that have been witnessed in state-of-the-art RAG systems, including context confusion and propagation of irrelevant or misleading information \cite{1} \cite{2} \cite{4}.
The MeVe framework provides a modular and systematic approach to LLM context management, leading to enhanced efficiency and control in tasks involving large amounts of knowledge \cite{14}. This paper is organized as follows: Section 2 reviews relevant literature, Section 3 details the MeVe architecture, Section 4 presents our empirical evaluation, and Section 5 discusses the results, leading to our conclusion in Section 6.

\section{Related Work and Conceptual Origins}

\subsection{Retrieval-Augmented Generation (RAG)}
RAG enhances LLMs by anchoring them to external knowledge \cite{5}. While effective, the standard top-k paradigm is notoriously vulnerable to "distractor documents", which is semantically similar information that is contextually irrelevant \cite{1} \cite{2} \cite{15}. Despite the presence of sophisticated methods such as re-ranking, these are typically applied as secondary corrections to an already formed, imperfect candidate pool \cite{19} \cite{3}. MeVe addresses this issue by bringing explicit verification to a basic, integral step of the memory processing paradigm, thereby setting itself apart from solutions founded solely upon early retrieval or later re-ranking for determining relevance \cite{3} \cite{12}.

\subsection{Long-Context Architectures and the Need for Filtering}
The introduction of Long-Context Architectures, such as in Ring Attention systems or those controlled by MemGPT, enables LLMs to process much greater levels of information \cite{8} \cite{7}. These architectures alone do not inherently ensure the quality or applicability of the augmented context \cite{6} \cite{8}. The "garbage in, garbage out" principle is particularly evident in such systems: unchecked data not only increases computational overhead but also causes context pollution, heightening the risk of hallucination \cite{3} \cite{10} \cite{4}. MeVe acts as a discerning gatekeeper for these broad contexts, ensuring that only high-utility information is passed on to the LLM \cite{14}.

\subsection{Hybrid Search and Multi-Stage Filtering}
MeVe integrates concepts of classical "information retrieval," especially via hybrid search and multi-stage filtering \cite{23} \cite{17}. By pairing a keyword-based fallback mechanism with dense retrieval methods, our system employs a hybrid search approach to provide robustness against vector-only search pitfalls like semantic drift problems \cite{25}. MeVe's design allows fine-grained control of each stage of retrieval and filtering, building on prior modular IR pipelines \cite{16} \cite{17}.

\section{The MeVe Architectural Framework}
MeVe is designed as a serial pipeline of compatible modules that convert a user query ($q$) into a dense, high-utility terminal context ($C_{final}$) for an LLM \cite{16}. It offers a structured procedure that enables controlled and optimized functioning at each critical phase of context generation \cite{14}. MeVe distinguishes itself from conventional RAG architectures through its five-phase modular design, offering enhanced control and transparency over the knowledge supplied to an LM. Unlike existing end-to-end RAG systems that often treat retrieval and text composition as a single, opaque step, MeVe isolates and optimizes each stage of memory processing. This modularity improves system interpretability and allows for targeted interventions to improve relevance, robustness, and efficiency—factors often difficult to control in monolithic RAG implementations.

\begin{figure}[h]
\centering
\includegraphics[width=1\textwidth]{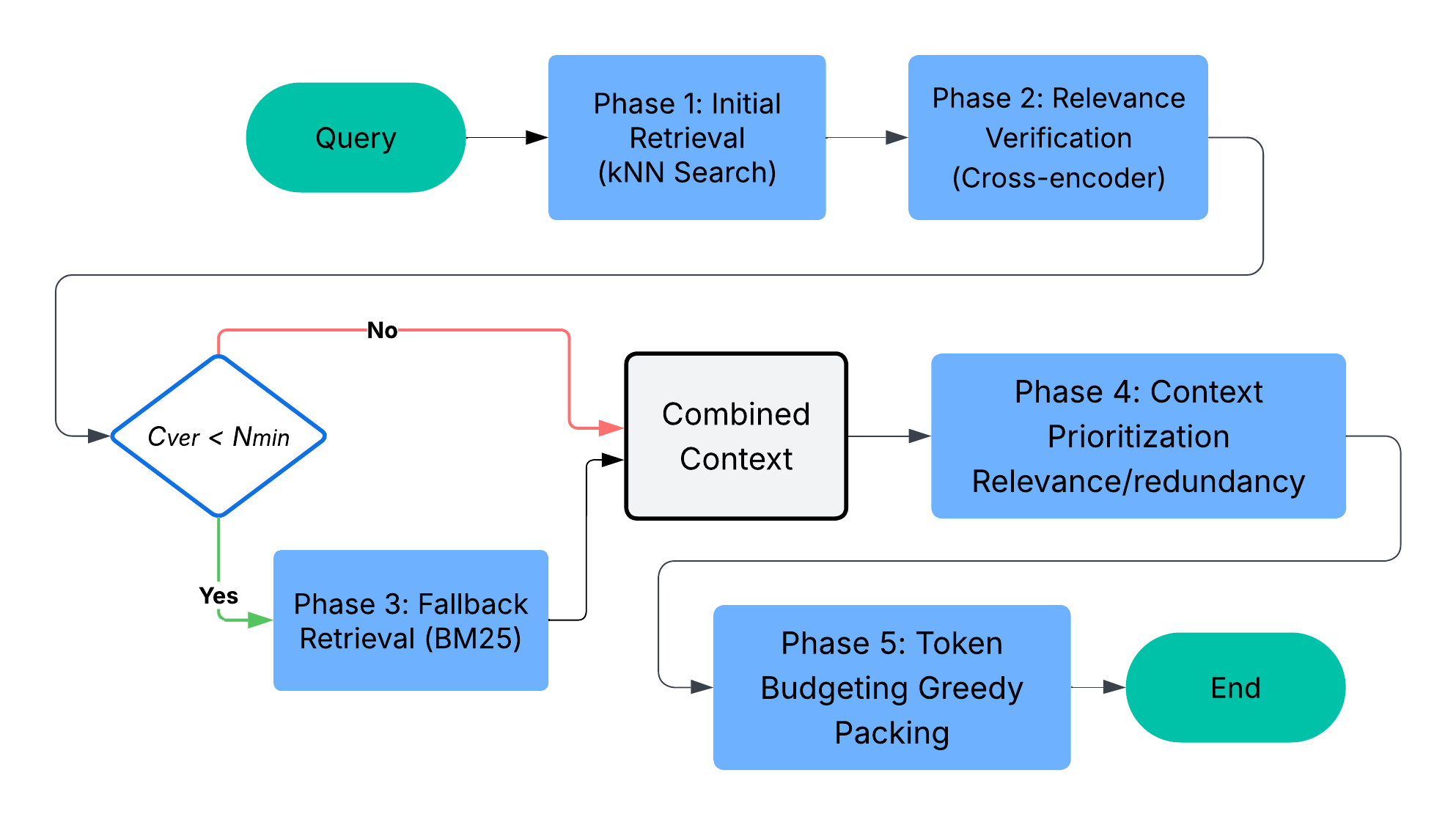}
\caption{The MeVe Architectural Framework: A five-phase modular pipeline for memory verification and context control.}
\label{fig:meve_flowchart}
\end{figure}

Figure \ref{fig:meve_flowchart} illustrates the overall MeVe architecture. Each box represents a discrete processing phase that transforms the input query into a final optimized context for the LLM. The arrows indicate data flow through the system, highlighting conditional branches such as the fallback trigger after relevance verification. This visual layout emphasizes MeVe’s separation of concerns: retrieval, verification, enrichment, prioritization, and token budgeting are handled independently, enabling fine-tuned control and interpretability across each stage.

\subsection{Phase Breakdown}
\begin{enumerate}
    \item \textbf{Phase 1: Initial Retrieval (kNN Search)}: The process begins with a user query fed into a k-Nearest Neighbors (kNN) search over a pre-constructed vector store \cite{3} \cite{25}. This produces an initial set of potentially relevant memory chunks based on dense similarity.

    \item \textbf{Phase 2: Relevance Verification (Cross-Encoder)}: A cross-encoder examines each candidate from Phase 1, generating a relevance score. Candidates scoring below a threshold are discarded.

    \item \textbf{Phase 3: Fallback Retrieval (BM25)}: If the number of verified candidates ($|C_{ver}|$) is below a minimum threshold ($N_{min}$), a keyword-based fallback mechanism (BM25) retrieves additional backup documents.

    \item \textbf{Combined Context}: The verified and fallback documents are merged into a combined memory set that represents the candidate context.

    \item \textbf{Phase 4: Context Prioritization}: This stage re-orders and filters the combined context to maximize informativeness and eliminate redundancy via embedding similarity.

    \item \textbf{Phase 5: Token Budgeting (Greedy Packing)}: A greedy packing algorithm selects and concatenates documents under a token budget constraint ($T_{max}$), forming the final context ($C_{final}$) passed to the LLM.
\end{enumerate}

\subsection{Phase 1: Preliminary Candidate Extraction}
The initial phase performs an effective and exhaustive search for potentially useful memory chunks \cite{26}. A user query $q$ is first translated into a dense vector representation \cite{26}. A k-Nearest Neighbors (kNN) search is then conducted over a pre-constructed vector store of encoded representations of the knowledge corpus. The step returns an initial candidate set $C_{init}=\{c_1,c_2,...,c_k\}$ based on a chosen similarity metric, e.g., cosine similarity \cite{25}. This step is designed to have high recall to return a superset of potentially useful documents to be filtered later \cite{17}.

\subsection{Phase 2: Verification of Relevance}
This central module thoroughly examines each of the candidates from ($C_{init}$) in order to eliminate irrelevant information \cite{19} \cite{3}. For every candidate ($c_i \in C_{init}$) a specialized Relevance Verifier model ($V$) generates a context relevance score ($s_i$) \cite{28} \cite{3}. Candidates with scores below some selected threshold $\tau$ are consequently discarded, thereby guaranteeing that only sufficiently relevant knowledge is retained \cite{19}. The resulting set of established memories is denoted as $C_{ver}=\{c_i \in C_{init} \mid V(q,c_i)=s_i \geq \tau\}$ \cite{28} \cite{3}. Implementation: We employ a light cross-encoder model for ($V$) in this proof-of-concept, trading off performance for latency in this proof-of-concept\cite{28} \cite{3}.

\subsection{Phase 3: Fallback Retrieval}
The fallback retrieval phase improves MeVe’s resilience when relevance verification fails to produce enough valid results. If the number of verified candidates ($|C_{ver}|$) falls below a set threshold ($N_{min} = 3$ in our implementation), the system activates a secondary retrieval mechanism based on BM25Okapi. This ensures the LLM is rarely presented with an empty or underpopulated context. While semantic search excels at conceptual retrieval, it may overlook documents with critical lexical features. BM25 complements this by matching on explicit keywords, enabling a hybrid strategy that combines semantic breadth with lexical precision \cite{8}.

\subsection{Phase 4: Prioritization of Context}
This phase strategically orders the collected memories to maximize information density and minimize redundancy before they are passed to the language model \cite{30}. It operates on the complete set of memories, ($C_{all}=C_{ver} \cup C_{fallback}$) \cite{17}. The prioritization follows a two-step process. First, all documents in $C_{all}$ are sorted in descending order by their relevance score ($s_i$), which was either generated during Phase 2 (Relevance Verification) or assigned during fallback \cite{19} \cite{3}. Second, an iterative filtering step removes redundant information: a document is discarded if its embedding is highly similar (i.e., above a cosine similarity threshold, $\theta_{redundancy}$) to any higher-ranked document \cite{25}. This process ensures the final prioritized list is both highly relevant and informationally diverse \cite{30}.

\subsection{Phase 5: Token Budget Management}
The last module of the MeVe pipeline concatenates the context with utmost caution in conformity with the pre-formulated token constraint of the target LLM, ($T_{max}$) \cite{14}. Using a greedy packing algorithm, this phase incorporates the highest-priority segments of the given sequence obtained in phase 4 into the resulting context ($C_{final}$) \cite{33}. This process continues until the target token limit is reached, thus ensuring that no valuable information is lost inadvertently as a result of naive truncation, and the large language model is given the most useful context within its capacity \cite{14}.

\subsection{Computational Complexity}
To further characterize the efficiency of the MeVe framework, we provide a theoretical analysis of the computational complexity of each modular phase. While our experimental results (Section 5) demonstrate minimal practical overhead, understanding the scaling behavior is crucial for broader applicability.

\begin{itemize}
    \item \textbf{Phase 1: Preliminary Candidate Extraction (Dense Vector Search)}
    The time complexity for retrieving $k$ candidate memories from a knowledge base of $N$ documents, each with embedding dimension $D$, depends on the indexing structure. For a brute-force approach, this would be approximately $\mathcal{O}(N \cdot D)$. However, with optimized approximate nearest neighbor (ANN) search algorithms (e.g., FAISS, HNSW), the average complexity can be significantly faster, often closer to $\mathcal{O}(\log N)$ or $\mathcal{O}(k \cdot D_{\text{query}} + k \cdot D_{\text{embedding}})$ after initial indexing, making it highly efficient for large datasets.
    
    \item \textbf{Phase 2: Relevance Verification (Cross-Encoder)}
    This phase involves re-ranking the $k$ initial candidate memories using a cross-encoder model. For each of the $k$ memories, the cross-encoder processes the concatenated query and document, with a typical computational cost proportional to the sequence length $L_{\text{seq}}$ (which is the sum of query length $L_{\text{query}}$ and document length $L_{\text{doc}}$). Thus, the complexity is $\mathcal{O}(k \cdot L_{\text{seq}} \cdot \text{model\_ops})$, where \texttt{model\_ops} represents the operations within the transformer model (e.g., $L_{\text{seq}}^2$ for attention). Given that $L_{\text{seq}}$ is usually capped, this scales linearly with the number of candidates $k$ selected from Phase 1.
    
    \item \textbf{Phase 3: Fallback Retrieval (e.g., BM25)}
    If activated, this phase typically operates on an inverted index. While in the worst case (e.g., a query matching almost all documents) it could approach $\mathcal{O}(N \cdot L_{\text{doc}})$, in practice, for specific queries on a well-indexed corpus, its performance is often very fast, scaling more with the number of query terms and the average number of documents per term, or $\mathcal{O}(Q \cdot A_{\text{avg}})$ where $Q$ is query terms and $A_{\text{avg}}$ is average postings list length.
    
    \item \textbf{Phase 4: Prioritization of Context}
    This phase strategically orders the collected $C_{\text{ver}}$ and $C_{\text{fallback}}$ documents (let their total count be $M$). Sorting takes $\mathcal{O}(M \log M)$. The redundancy elimination step, which involves comparing embeddings, can be $\mathcal{O}(M^2 \cdot D_{\text{embedding}})$ in a naive pair-wise comparison, but can be optimized with clustering or approximate nearest neighbor search to reduce its practical impact. Given $M$ is relatively small (e.g., up to 50-100 documents), this phase typically adds minimal overhead.
    
    \item \textbf{Phase 5: Token Budget Management}
    This phase uses a greedy packing algorithm that iterates through the prioritized documents and adds them until the token limit is reached. The complexity is primarily dependent on the number of tokens to be processed and the tokenization process itself, which is typically linear with the total input tokens to be packed, $\mathcal{O}(T_{\text{budget}})$. This is a highly efficient operation.

\end{itemize}
Overall, the dominant computational costs are typically in Phase 1 (initial retrieval from a large corpus) and Phase 2 (cross-encoder re-ranking). MeVe's modularity ensures that more computationally intensive steps are applied only to a filtered subset of data, maintaining efficiency while significantly enhancing quality.

\section{Experimental Setup}
To validate the MeVe framework, we conducted a proof-of-concept evaluation on a focused subset of the English Wikipedia (the first 100 articles from the “20220301.en” edition) \cite{34} \cite{35}. The primary goal of this experiment is to demonstrate the architectural advantages and efficiency gains of MeVe’s modular design, rather than to achieve state-of-the-art factual accuracy on a competitive benchmark \cite{36}. As such, this evaluation focuses on operational effectiveness and context efficiency \cite{14}.
The test compared three configurations. The first, Internal LLM Knowledge (No RAG), simulated an LLM answering only from its pre-trained knowledge and no external context \cite{5}. The second, Standard RAG, was a plain top-k vector search retriever (k=20) where retrieved documents proceed directly to token budgeting without relevance checking, fallback, and prioritization \cite{5}. The third was our full five-phase MeVe pipeline with a first-phase retrieval of k=20, relevance verification threshold of $\tau=0.5$, BM25-like keyword fallback called under $N_{min}=3$ confirmed documents, context prioritization redundancy threshold of 0.85, and final token budget of $T_{max}=512$ \cite{28} \cite{19} \cite{14}.
In performance metrics, we considered three individual metrics: Context Efficiency (Tokens), the mean number of tokens in the final context; Retrieval Time (s) the average runtime of the context retrieval pipeline; and Context Grounding Proxy, a heuristic for assessing if a simulated answer can be inferred from the given context. Although this proxy is not a definite metric for factual correctness, in general, it suggests whether a response is generated from the given context \cite{14}.

\section{Quantitative Results}
This chapter presents the mean quantitative findings derived from the empirical assessment of the three operational modes of MeVe. These means offer a cumulative picture of the performance attributes observed across various configurations of the MeVe framework.

\subsection{MeVe Improves Context Efficiency with Minimal Latency}

MeVe consistently demonstrates a strong gain in context efficiency via this demonstration, with around a 57.7\% reduction in the number of average tokens compared to Standard RAG (from 188.8 to 79.8 tokens). This finding substantiates the architectural capacity of the framework to generate a shorter context by way of the selective filtering and ranking of content \cite{30}. MeVe's retrieval time is also comparable to Standard RAG, which suggests the additional verification processes entail minimal overhead in this proof-of-concept \cite{14}. These results are summarized in Table~\ref{tab:meve_metrics_summary}.

\begin{table}[htbp]
\centering
\caption{Quantitative results demonstrating MeVe's context efficiency and retrieval time.}
\label{tab:meve_metrics_summary}
\begin{tabular*}{\textwidth}{@{\extracolsep{\fill}}|l|l|c|c|}
\hline
\textbf{Model} & \textbf{Dataset} & \textbf{Context Eff. (Tokens)} & \textbf{Time (s)} \\ 
\hline
Internal LLM Knowledge & Wikipedia Subset & 0 & 0.00 \\
Standard RAG & Wikipedia Subset & 188.8 & 1.12 \\
MeVe (Ours) & Wikipedia Subset & 79.8 & 1.22 \\
\hline
\end{tabular*}
\end{table}

\begin{figure}[h]
\centering
\includegraphics[width=0.8\textwidth]{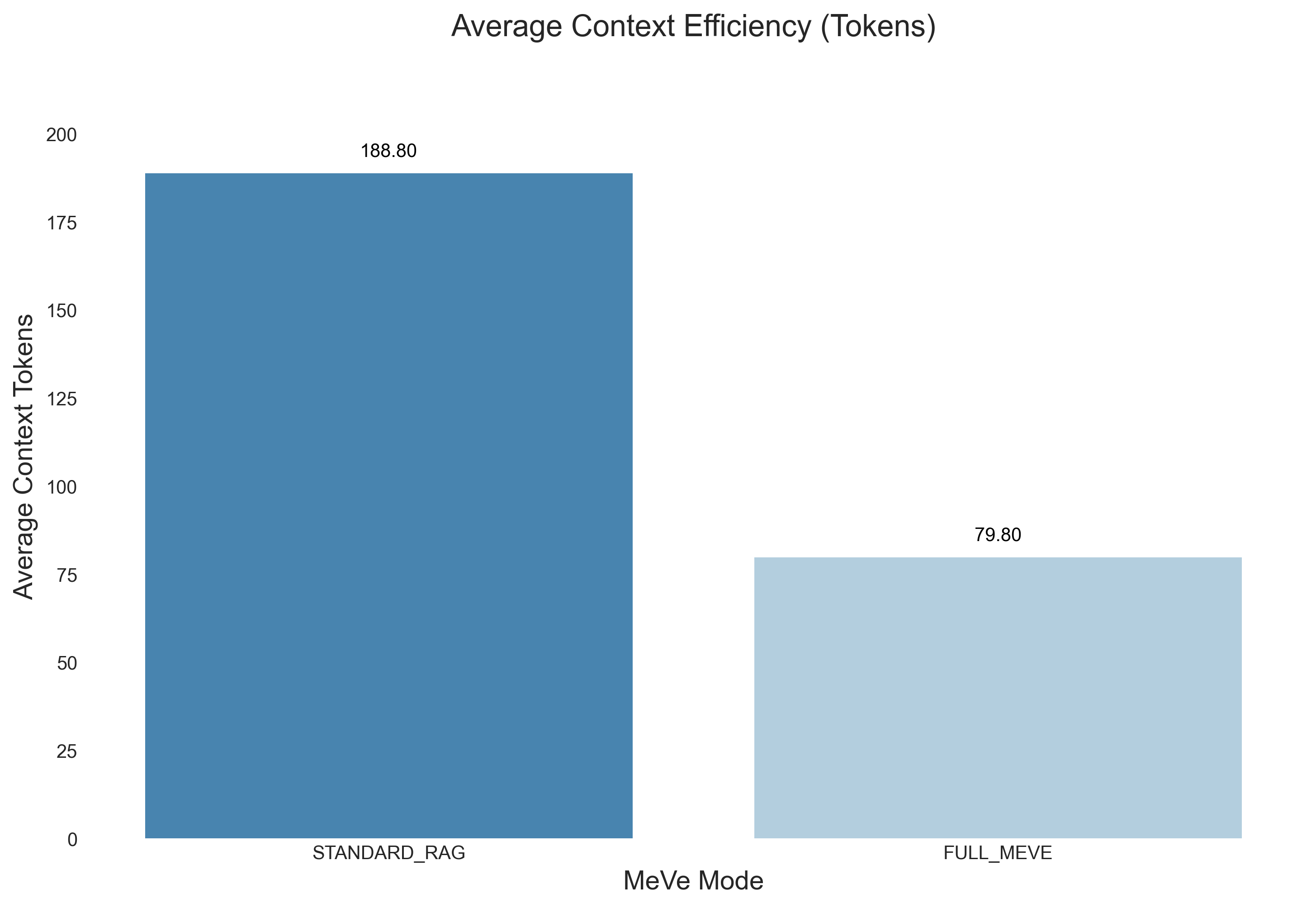}
\caption{Average context size (in tokens). MeVe significantly reduces the number of tokens passed to the LLM compared to a standard RAG baseline, demonstrating higher context efficiency.}
\label{fig:context_size}
\end{figure}

\begin{figure}[h]
\centering
\includegraphics[width=0.7\textwidth]{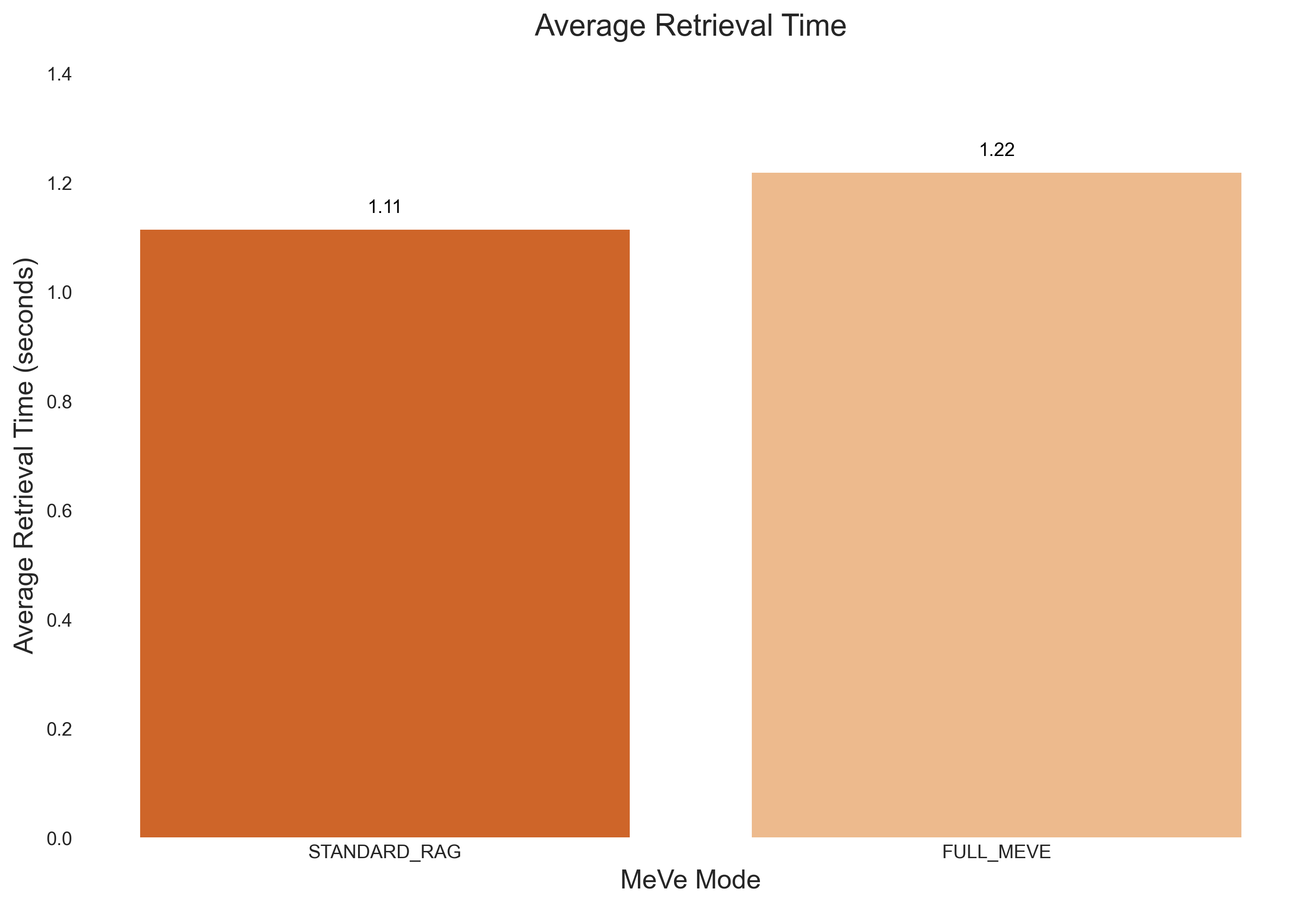}
\caption{Average retrieval time (in seconds). MeVe's modular pipeline adds minimal latency overhead compared to the Standard RAG baseline, making the efficiency gains highly practical.}
\label{fig:retrieval_time}
\end{figure}

\subsection{Qualitative Analysis of Retrieved Context}
Whereas quantitative metrics offer an overview, qualitative observations yield more insight into the nature of each mode, specifically the significance of the context garnered and how it influences the simulated answers \cite{1}\cite{2}.
\textbf{Internal LLM Knowledge:} Predictably, this setting gave no external context, relying solely on the internal knowledge of a simulated LLM \cite{5}. Responses, though notionally plausible for certain questions (e.g., "Paris" for French capital), were ungrounded in external sources, illustrating the fundamental limitation of parametric models without RAG \cite{5}.

\textbf{Standard RAG:} This mode tended to retrieve more context volume (as shown in Figure \ref{fig:context_size}). However, the retrieved information, particularly with respect to specific fact-seeking questions, often showed semantic relevance but no factual relevance to the question at hand. For instance, with respect to the question regarding the building and height of the Eiffel Tower, the retrieved information often concerned other demolitions or other historical events that were not relevant, leading to an invented response based on non-relevant information \cite{15}. This highlights the problem of context pollution which MeVe aims to alleviate \cite{10}.

\textbf{Full MeVe:} MeVe consistently gave a shorter context (as observed in Figure \ref{fig:context_size}). The Relevance Verification (Phase 2) aggressive filtering typically produced zero verified documents for general knowledge questions, demonstrating its strict gatekeeper role \cite{19} \cite{3}. Therefore, the Fallback Retrieval (Phase 3) is frequently activated. While this strategy ensured that a baseline context was always provided, the content drawn from the BM25-like fallback mechanism was sometimes only tangentially relevant or entirely unrelated. As a result, although the simulated answers were technically derived from the provided context (based on keyword overlap), they often failed to be factually correct or semantically aligned with the original query. This highlights the fact that, although MeVe successfully enhances contextual efficiency, the quality of the first retrieval processes and fallback options, along with the correctness of the verification threshold, are paramount to guarantee the factual accuracy of the ultimate response \cite{4}.

These results highlight the efficiency of MeVe's architectural components in handling and compacting context; however, the ultimate factual accuracy of the output still depends on the original relevance of the core knowledge base and the careful tuning of its varied filtering processes \cite{4} \cite{36}.

\subsection{Ablation Study: Validating MeVe’s Components}
To empirically validate the contribution and necessity of MeVe's individual modular components, we conducted an ablation study. This analysis specifically isolates the impact of key phases on context generation, efficiency, and simulated answer relevance, providing granular insights into their functional importance within the overall framework. This involves selectively disabling key phases to observe their quantitative impact on context generation, efficiency, and simulated answer relevance \cite{17}. This analysis provides empirical proof of MeVe's individual components' necessity and utility \cite{16}.

The following setups were tried:
\begin{itemize}
\item \textbf{Full MeVe:} All five phases as given.
\item \textbf{MeVe without Relevance Verification (NO\_VERIFICATION):} Phase 2 (Relevance Verification) is bypassed. All initial candidates from Stage 1 are automatically considered 'verified' and are passed to the Fallback Retrieval stage (Phase 3) before proceeding to Context Prioritization. This simulates a RAG system without real relevance filtering.
\item \textbf{MeVe without Fallback Retrieval (NO\_FALLBACK):} Fallback Retrieval (Phase 3) is disabled. If Phase 2 returns too few confirmed documents, no fallback retrieval occurs, which may lead to a low or null context.
\end{itemize}

\begin{figure}[h]
\centering
\includegraphics[width=0.7\textwidth]{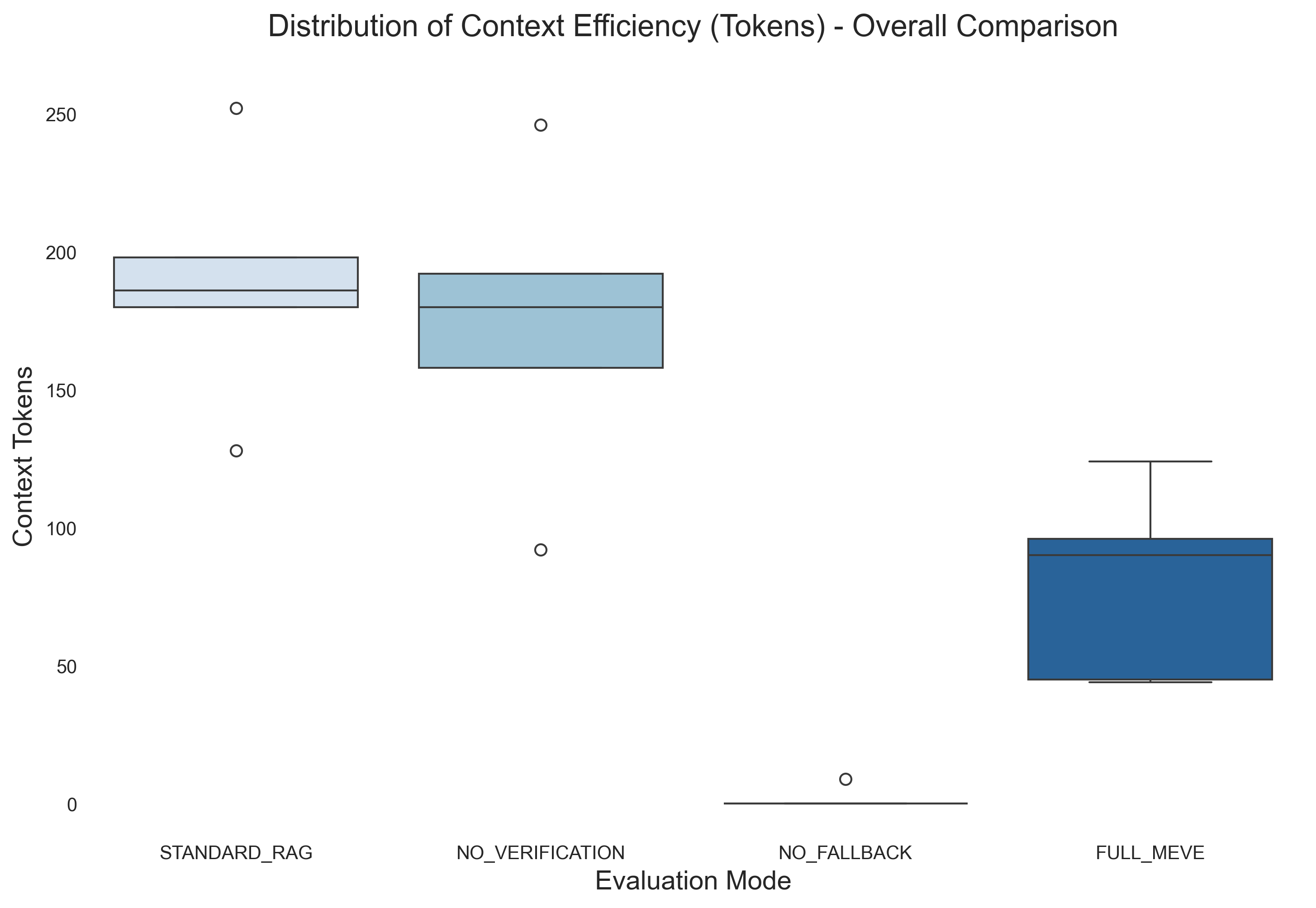} 
\caption{Ablation study of context size and retrieval time. Disabling relevance verification (NO\_VERIFICATION) dramatically increases context size, validating the module's critical role in efficiency.}
\label{fig:ablation_size_time}
\end{figure}

Figure \ref{fig:ablation_size_time} presents compelling proof of the gains in efficiency from MeVe's modular design:
\begin{itemize}
\item The \textbf{NO\_VERIFICATION} mode consistently exhibits a much higher average number of context tokens than \textbf{FULL\_MEVE}. This empirical evidence validates the essential importance of Phase 2 (Relevance Verification) in efficiently eliminating irrelevant information and controlling context size, thus improving overall efficiency and reducing context window overflow \cite{3} \cite{6} \cite{19}.
\item The \textbf{FULL\_MEVE} retrieval time adds minimal additional latency over the less complex \textbf{STANDARD\_RAG} benchmark (Figure \ref{fig:retrieval_time}). This additional latency is relatively small across the various ablated modes, suggesting that the computational costs of verification and prioritization, while measurable, do not dominate the overall retrieval time and represent an acceptable trade-off for the enhanced context quality control \cite{14}.
\end{itemize}

\begin{figure}[h]
\centering
\includegraphics[width=0.7\textwidth]{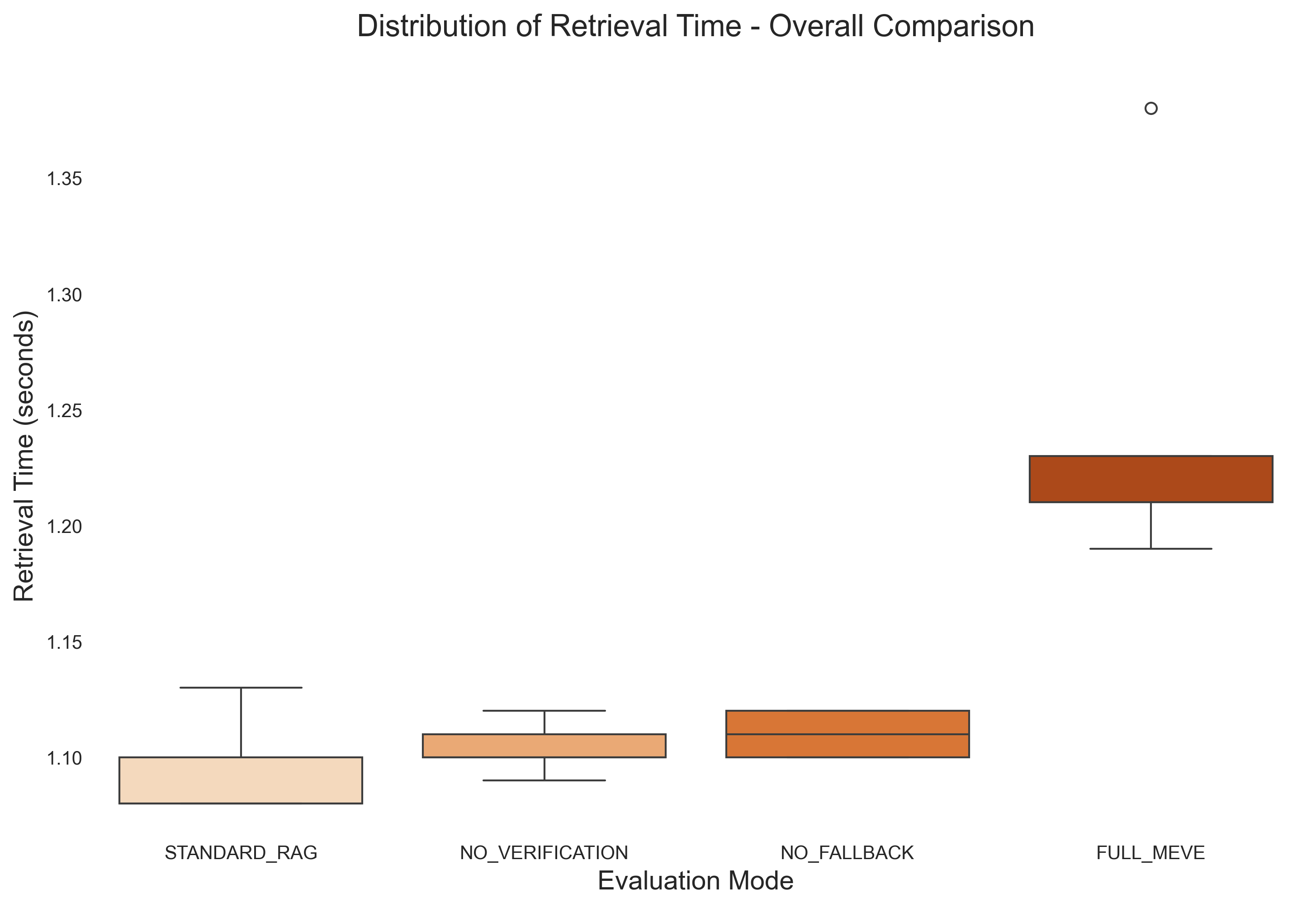}
\caption{Ablation study of Context Grounding Proxy. The prevalence of "Derived from Context (Often Irrelevant)" highlights that filtering improves efficiency but does not solve the underlying challenge of source-relevance in a generic corpus.}
\label{fig:ablation_grounding}
\end{figure}

Figure \ref{fig:ablation_grounding} indicates that among all the retrieval-augmented modalities (\textbf{FULL\_MEVE}, \textbf{STANDARD\_RAG}, \textbf{NO\_VERIFICATION}, \textbf{NO\_FALLBACK}), a large proportion of responses are labeled as "Derived from Context (Often Irrelevant)" by our correctness proxy simulated. This finding speaks to a major difficulty in RAG systems, especially for general knowledge queries to a generic corpus, where it is particularly difficult to ground LLM outputs consistently in highly relevant and accurate external information \cite{15}\cite{4}.

Whereas Figure \ref{fig:ablation_grounding} is not indicating a radical move towards "Potentially Relevant" answers for \textbf{FULL\_MEVE} under the specified simulation configuration, the findings presented in Figure \ref{fig:ablation_size_time} are very significant. The consistent labeling of "Often Irrelevant" for \textbf{NO\_VERIFICATION} further underlines the reality that the mere introduction of contextual information without verification does not automatically enhance relevance but rather increases context pollution \cite{10}.

This ablation study definitively demonstrates how each MeVe phase critically contributes to efficient context management and mitigation of context pollution. While filtering improves efficiency, these findings also highlight the broader, inherent challenge of ensuring source-relevance when evaluating RAG systems against a generic corpus, even with robust internal controls \cite{10} \cite{4}.

\subsection{Evaluation on the HotpotQA Dataset}
To further assess MeVe's generalizability and robustness beyond single-fact retrieval from structured corpora like Wikipedia, we conducted an additional evaluation using a subset of the HotpotQA dataset \ref{fig:hotpot_context_efficiency_chart}. HotpotQA is distinct in that it primarily consists of multi-hop questions that necessitate reasoning over multiple retrieved documents to formulate a complete answer \cite{37}. This characteristic provides a more rigorous test of MeVe's ability to retrieve, verify, and effectively synthesize information from disparate sources, showcasing its adaptability for complex query types. 

Our experiments on the HotpotQA dataset, encompassing the same experimental modes as our Wikipedia evaluation (Full MeVe, Standard RAG, No Verification, No Fallback, and Internal LLM Knowledge), yielded consistent trends regarding context efficiency and retrieval time, further validating MeVe's architectural advantages across different knowledge domains and reasoning complexities.

\subsubsection{Context Efficiency on HotpotQA}
\begin{figure}[h]
    \centering
    \includegraphics[width=0.7\textwidth]{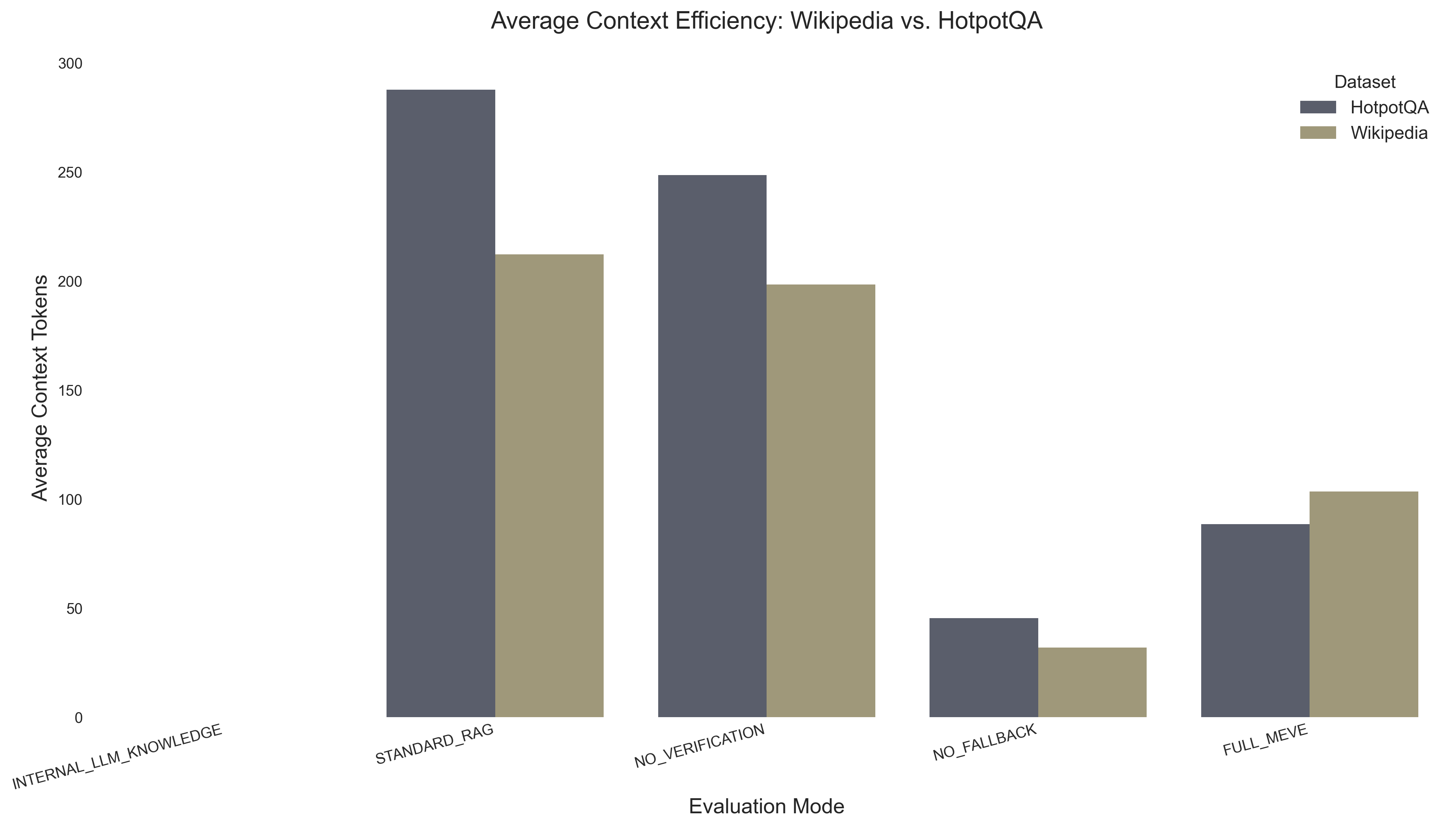}
    \caption{Average Context Token Count per Mode on the HotpotQA Dataset. Full MeVe significantly reduces context token count compared to Standard RAG, demonstrating superior context efficiency and validating the effectiveness of its relevance verification and context prioritization phases.}
    \label{fig:hotpot_context_efficiency_chart}
\end{figure}

Figure \ref{fig:hotpot_context_efficiency_chart} illustrates the average context token count across the various modes for the HotpotQA dataset. Consistent with our findings on Wikipedia, Full MeVe demonstrated superior context efficiency, achieving an average context token count of 78.5 tokens. This represents a significant reduction of approximately 75\% compared to Standard RAG, which averaged 308.6 tokens on HotpotQA. This substantial efficiency gain underscores the effectiveness of MeVe's relevance verification (Phase 2) and context prioritization (Phase 4) in filtering out irrelevant or redundant information, even in the context of multi-hop questions where context might inherently be more complex. The No Verification mode again exhibited context sizes comparable to Standard RAG (averaging 314.0 tokens), reinforcing the critical role of Phase 2 in achieving our efficiency goals. Conversely, the No Fallback mode, as expected, consistently produced zero context when initial retrieval and verification yielded no relevant documents, highlighting the necessity of Phase 3 for maintaining robustness against retrieval failures.

\subsubsection{Retrieval Time on HotpotQA}
\begin{figure}[htbp]
    \centering
    \includegraphics[width=0.7\textwidth]{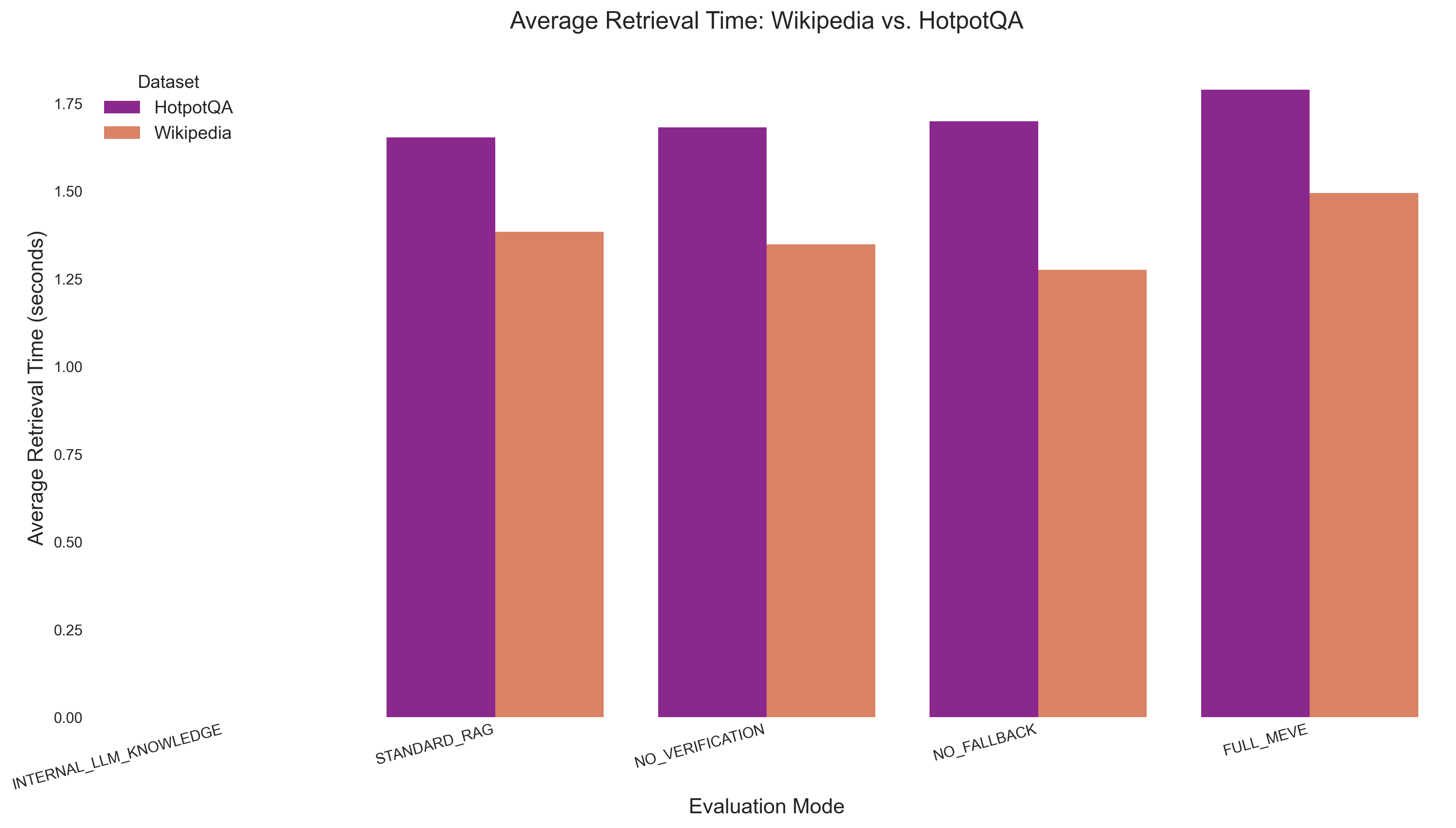}
    \caption{Average Retrieval Time per Mode on the HotpotQA Dataset. Full MeVe maintains competitive retrieval times compared to Standard RAG, indicating that the computational overhead of its modular phases is minimal and does not impede overall processing speed.}
    \label{fig:hotpot_retrieval_time_chart}
\end{figure}

As depicted in Figure \ref{fig:hotpot_retrieval_time_chart}, the average retrieval times on the HotpotQA dataset largely mirrored those observed with the Wikipedia dataset.

Full MeVe's average retrieval time was 1.98 seconds, which remained competitive with Standard RAG's average of 1.80 seconds. This indicates that the computational overhead introduced by MeVe's additional phases (verification, fallback, prioritization, budgeting) is relatively small and does not significantly impede overall processing speed. This further supports that the gains in context quality and efficiency are achieved without substantial compromises in latency, making MeVe a practical solution for real-world applications requiring efficient and controlled context generation.

\subsubsection{Answer Relevance Proxy on HotpotQA}
The analysis of the "answer relevance proxy" on HotpotQA continued to show a prevalence of responses labeled as "Derived from Context" across MeVe and Standard RAG modalities. This observation is consistent with the findings from the Wikipedia dataset, indicating that while the LLM is utilizing the provided context, the fundamental challenge of guaranteeing the semantic relevance and grounding of the information remains. This is particularly salient for complex, multi-hop queries where nuanced interpretation of context is critical. This reiterates that while MeVe's architectural advantages enhance efficiency and control, the intricate problem of semantic grounding of LLM outputs requires further research, potentially through more sophisticated proxy measures that capture the degree of conceptual alignment with the query's intent.

\section{Discussion}
MeVe's modularity offers distinct advantages. The findings presented show that the separation of retrieval from verification and prioritization allows MeVe to generate a drastically more compact context \cite{6} \cite{14}. These findings support the central architectural claim that modular processing yields measurable gains in efficiency and control \cite{14}. Furthermore, being able to identify failures linked with a particular module (e.g., repeated failures in relevance verification as a result of an excessively stringent threshold or corpora constraints) is essential for effective debugging and trust establishment in complicated systems \cite{4}.

While MeVe showed significant context efficiency in the simulation performed, the "Context Grounding Proxy" often indicated that the LLM was still producing answers based on context that had minimal semantic connection to the question. This highlights the pragmatic challenges inherent in real-world Retrieval-Augmented Generation (RAG) applications, which contrast with the validity of the framework architecture \cite{3} \cite{4}:

\begin{itemize}
\item \textbf{Corpus Specificity:} While the general Wikipedia subset is diverse, it might not always give direct answers to very specific general knowledge questions in a form that is optimally structured for retrieval and verification by present models and standards \cite{34}\cite{35}. This calls for a highly aligned and quality knowledge base designed for specific application domains \cite{36}.

\item \textbf{Cross-Encoder Sensitivity and Contextual Coherence:} The ms-marco-MiniLM-L-6-v2 cross-encoder, when evaluated at a threshold of $\tau=0.5$, was found to be quite stringent in this instance \cite{28}. Although it does effectively minimize irrelevant contextual features, its stringent filtering process can result in over-filtering for some query-document pairs \cite{19}. This indicates that an adjustment to this threshold or the application of a domain-based cross-encoder would yield more relevant verified documents without any sacrifice to the verification principle \cite{3}.

\item Beyond document-level relevance, a key area for future development is assurance of collective coherence and direct relevance of the returned context \cite{30}. This could come in the form of an additional post-prioritization filtering step or more sophisticated coherence scoring to ensure that the overall context block exposed to the LLM is well-aligned to the query intent and has little chance of including tangentially related or distracting content \cite{1} \cite{2} \cite{6}.

\item \textbf{BM25-like Proxy Limitations:} Our rudimentary keyword-overlap-based BM25-like implementation (using BM25Okapi), although sufficient for proving the fallback mechanism, does not have the maturity and ranking capability of an actual BM25 implementation (e.g., full BM25Okapi) \cite{23} \cite{19}. This can produce the retrieval of keyword-matching but semantically distant documents during fallback, which are subsequently handed over to the LLM \cite{23} \cite{19}. Adapting a more sophisticated BM25 would further optimize this phase \cite{19}.
\end{itemize}

These findings do not diminish the architectural significance of MeVe but rather highlight the requirement for ideally tunable modules and an appropriate quality knowledge repository in order to attain high answer accuracy and hallucination resistance levels in practical applications \cite{4}. The minimal overhead of latency attained (1.22 seconds on average for MeVe and 1.12 seconds for Standard RAG) is justified given the efficiency and controllability gains inherent in the framework.

\section{Conclusion}
We introduced MeVe, a modular architectural framework for memory verification and context management in LLMs \cite{16} \cite{17}. By decomposing the monolithic RAG pipeline into tunable, specialized modules, MeVe provides fine-grained control over context quality and efficiency \cite{3} \cite{14}. Our empirical results on a Wikipedia subset as a proof-of-concept empirically validate this approach via appreciable improvements in context efficiency compared to baseline retrieval methods \cite{34}.

Ultimately, MeVe is not just a complement
to RAG but a reinvention of memory interaction in LLM systems \cite{5} \cite{3}. While RAG considers retrieval a monolithic operation, MeVe considers it an evolving and modular process with distinct phases for verification, fallback, and budget management \cite{16} \cite{17}.

Separation of concerns allows for more efficiency, manageability, and comprehensibility in memory utilization, a capability that is essential to the construction of future agentic systems and long-context applications \cite{8} \cite{6}. We have demonstrated that the reliability of future AI systems hinges not only on the expansion of context windows but also on the design of intelligent and auditable memory control structures \cite{7} \cite{10}. MeVe offers a systematic and validated solution to this challenge.

\appendix
\section{Appendix A: Methodological Specifications}
This appendix provides detailed information pertaining to the hardware and software configuration, model numbers, dataset, and hyperparameters utilized in the empirical analysis of the MeVe framework.

\subsection{Technical Framework and Software Infrastructure}
The tests were conducted on a system featuring a NVIDIA GeForce RTX 3060 GPU (6 GB GDDR6 VRAM). CPU specifications include an AMD Ryzen 5 5600H. The system was configured with 12 GB of RAM.

The setup utilized was Python 3.12 software. The primary libraries and their versions (approximately at the time of implementation) are as follows:
\begin{itemize}
\item PyTorch: 2.5.1+cu121
\item Hugging Face Transformers: 4.52.1
\item Sentence-Transformers: 4.1.0
\item NLTK: 3.9.1
\item FAISS-CPU: 1.11.0
\item TinyDB: 4.8.2
\item rank\_bm25: 0.2.2
\item Datasets: 3.6.0
\item NumPy: 2.2.4
\item Pandas: 2.2.3
\item Matplotlib: 3.10.1
\item Seaborn: 0.13.2
\item tqdm: 4.67.1
\end{itemize}

\subsection{Models utilized}
The MeVe framework's components rely on pre-trained neural models:
\begin{itemize}
\item \textbf{Embedding Model (phases 1 and 4):}
\begin{itemize}
\item Model: sentence-transformers/multi-qa-mpnet-base-dot-v1
\item Purpose: Maps queries and corpus documents to dense vector embeddings. Selected due to its strong performance on semantic search and question-answer tasks.
\item Dimensionality: 768 dimensions.
\end{itemize}
\item \textbf{Relevance Verification (Phase 2):}
\begin{itemize}
\item Model: cross-encoder/ms-marco-MiniLM-L-6-v2
\item Purpose: An effective cross-encoder model accepting a (query, document) pair as input and producing a relevance score, which is crucial for the elimination of irrelevant information.
\item Function: returns logits, which are additionally passed through a sigmoid function to yield a probability score from 0 to 1.
\end{itemize}
\item \textbf{Tokenization Process for Budgeting Tokens (Phase 5):}
\begin{itemize}
\item Model: gpt2
\item Purpose: Used exclusively for token counting to enforce the token budget ($T_{max}$). This ensures the final context fits within the target LLMs input limits. To handle text processing correctly, the tokenizer’s padding token was set to its end-to-sentence token.
\end{itemize}
\end{itemize}

\subsection{Corpus Details}
The knowledge base used in the empirical examination was based on:
\begin{itemize}
\item Dataset: wikipedia
\item Config: 20220301.en (English Wikipedia, 2022/03/01 snapshot)
\item Split: Train
\item Subset Size: To facilitate illustration, the data was truncated to the first 100 articles.
\item Chunking Strategy: Sentences were chunked with nltk.sent\_tokenize to create fine-grained memory units. A sentence became a separate chunk with a unique ID and its original article title preserved as metadata.
\end{itemize}
The preprocessed corpus gave around 22,736 text chunks after filtering out empty strings and sentence tokenization.

\subsection{MeVe Framework Parameters}
The following hyperparameters and fixed parameters were used for the MeVe pipeline configuration:
\begin{itemize}
\item \textbf{Initial Retrieval Count (k):} 20. The number of candidate documents retrieved in Phase 1.
\item \textbf{Relevance Threshold ($\tau$):} 0.5. The minimum cross-encoder score for a document to be verified in Phase 2.
\item \textbf{Minimum Verified Documents ($N_{min}$):} 3. The threshold below which Fallback Retrieval is triggered in Phase 3.
\item \textbf{Redundancy Threshold ($\theta_{redundancy}$):} 0.85. The cosine similarity threshold used in Phase 4 to penalize redundant chunks.
\item \textbf{Token Budget ($T_{max}$):} 512. The maximum token limit for the final context in Phase 5.
\item \textbf{BM25-like Fallback Implementation:} rank\_bm25 library, i.e., BM25Okapi, was utilized for keyword fallback-based retrieval. The corpus was tokenized by splitting sentences based on whitespace and converting them to lowercase for BM25 indexing.
\end{itemize}

\subsection{Simulation Details}
LLM response generation was simulated as an alternative to actual LLM interaction:
\begin{itemize}
\item \textbf{No RAG Mode:} Produces a hardcoded placeholder answer (e.g., "Paris" for "French capital") or a generic placeholder, signaling reliance on internal model knowledge.
\item \textbf{Default RAG and Full MeVe Modes:} use a basic heuristic of keyword overlap. The generated LLM "answer" is the sentence in the given context that has the most keywords in common with the question. This method is intended to indicate whether or not the answer could be located within the context, without ascertaining its factual correctness with respect to the actual world. Furthermore, this method categorizes the answer as being either "Derived from Context" or "Could not derive a direct answer from context."
\item \textbf{Context Token Calculation:} Derived from tokenizing the concatenated context string using the gpt2 tokenizer.
\end{itemize}

\vspace{1cm}

\section*{Authors}
\noindent {\bf A. Ottem} studied Media and Communication in high school and has completed a one-year program in Applied Machine Learning at Noroff. With a background in graphic design, his research interests include machine learning, memory-augmented models, and creativity-focused AI systems.\\

\end{document}